\documentclass[letterpaper, 10 pt, conference]{ieeeconf}  

\IEEEoverridecommandlockouts                              

\usepackage{amsmath}
\usepackage{amssymb}
\usepackage{mathtools}

\usepackage[T1]{fontenc}
\usepackage{multirow}
\usepackage{caption}
\usepackage{multicol}
\usepackage{siunitx}
\usepackage{booktabs}
\usepackage{subcaption}
\usepackage{wrapfig}
\usepackage[section]{placeins}
\usepackage{graphicx}
\usepackage{xcolor}
\definecolor{lightblue}{RGB}{100,160,235}
\usepackage[colorlinks=true,urlcolor=lightblue,linkcolor=black,citecolor=black]{hyperref}
\usepackage{cuted}
\usepackage[colorinlistoftodos]{todonotes}
\def\framename{LeanGate} 

\title{\LARGE \bf
Accelerating Transformer-Based \\ Monocular SLAM via Geometric Utility Scoring
}
\author{%
Xinmiao Xiong$^{1*}$, Bangya Liu$^{1*}$, Hao Wang$^{2}$, Dayou Li$^{2}$, Nuo Chen$^{2}$, \\
Andrew Feng$^{3}$, Mingyu Ding$^{4}$, Suman Banerjee$^{1}$, Yang Zhou$^{2}$, Zhiwen Fan$^{2\dagger}$\\
$^{1}$UW--Madison \quad $^{2}$Texas A\&M  \quad $^{3}$USC \quad $^{4}$UNC Chapel Hill%
\thanks{$^{\dagger}$Corresponding author: zhiwenfan@tamu.edu.}%
\thanks{$^{*}$Equal contribution.}%
}

\IEEEaftertitletext{%
\vspace{-5mm}
\begin{center}
  \captionsetup{type=figure}
  \includegraphics[width=0.9\textwidth]{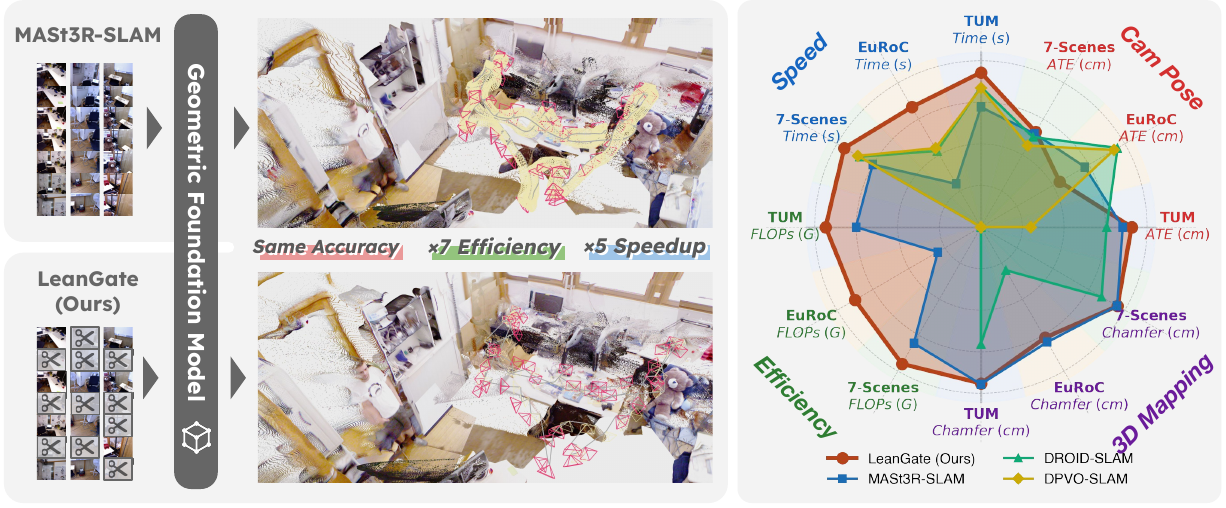}
  \caption{We present~\framename, a geometry-aware lightweight frame gating network that bypasses over $90\%$ of input frames in dense streaming (red: keyframes; yellow: redundant frames), while preserving the fidelity of mapping, tracking, and camera pose estimation. 
  We report total GPU FLOPs as the measure of efficiency, and end-to-end runtime as the measure of speedup. Since the reduction in time over the full scene is typically smaller than the overall FLOPs reduction, we report both metrics.}
  \vspace{-0.3cm}
  \label{fig:teaser}
\end{center}
\vspace{0.3\baselineskip}%
}
\begin{document}
\maketitle
\thispagestyle{empty}%
\pagestyle{empty}%

\begin{abstract}
Geometric Foundation Models (GFMs) have recently advanced monocular SLAM by providing robust, calibration-free 3D priors. However, deploying these models on dense video streams introduces significant computational redundancy. Current GFM-based SLAM systems typically rely on post-hoc keyframe selection. Because of this, they must perform expensive dense geometric decoding simply to determine if a frame contains novel geometry, resulting in late rejection and wasted computation. To mitigate this inefficiency, we propose~\framename, a lightweight feed-forward frame-gating network. \framename~predicts a geometric utility score to assess a frame's mapping value prior to the heavy GFM feature extraction and matching stages. By serving as a predictive plug-and-play module, our approach bypasses over 90\% of redundant frames. Evaluations on standard SLAM benchmarks demonstrate that~\framename reduces tracking FLOPs by more than 85\% and achieves a 5$\times$ end-to-end throughput speedup. Furthermore, it maintains the tracking and mapping accuracy of dense baselines. Project page: \href{https://lean-gate.github.io/}{https://lean-gate.github.io/}

\end{abstract}

\section{Introduction}
\label{sec:intro}
Simultaneous Localization and Mapping (SLAM) from a monocular camera serves as the spatial perception engine for modern autonomous robotics~\cite{cadena2017past, marchand2015pose} and augmented reality (AR) applications \cite{schops2014semi}. Traditional monocular SLAM architectures rely on multi-stage pipelines that process dense video streams, extract sparse features, and enforce handcrafted geometric constraints to jointly estimate camera poses and 3D map structures via backend Bundle Adjustment (BA) \cite{klein2007parallel, mur2015orb,mur2017orb,furukawa2015multi, engel2015large,schonberger2016pixelwise, yao2018mvsnet}. Conversely, dense SLAM frameworks bypass intermediate sparse extraction to optimize photometric consistency or predict dense surfaces directly from raw pixels \cite{newcombe2011dtam, zhou2018deeptam, bloesch2018codeslam, czarnowski2020deepfactors, teed2021droid}. Both paradigms, however, demand meticulous parameter tuning and frequently degrade under challenging conditions such as weak texture, rapid sensor motion, or dynamic illumination, restricting their robustness across diverse environments.

Recently, 3D Geometric Foundation Models (GFMs) have emerged as robust, data-driven alternatives for visual perception. Models such as DUSt3R \cite{wang2024dust3r}, MASt3R \cite{leroy2024grounding}, and VGGT \cite{wang2025vggt} take uncalibrated images and regress dense 3D representations, such as pointmaps, in a single forward pass. By learning multi-view geometric priors from massive datasets, GFMs bypass the fragility of traditional feature matching, resolving ill-posed reconstructions in textureless regions. Consequently, they offer highly stable front-ends for tracking and mapping in systems like MASt3R-SfM \cite{duisterhof2025mast3r} and MASt3R-SLAM \cite{murai2024_mast3rslam}. However, deploying these models on resource-constrained platforms remains challenging. Processing dense video streams (e.g., 30 FPS) at full resolution introduces severe computational redundancy, precluding real-time performance.

This computational redundancy primarily stems from an operational mismatch between how GFMs are trained and how they are deployed in SLAM. GFMs are inherently designed to recover geometry from sparse views with large baselines \cite{fan2024instantsplat}. Yet, current GFM-based SLAM systems process dense temporal streams, incurring heavy encoding and decoding costs on nearly every frame. For instance, in MASt3R-SLAM, dense feature extraction accounts for over 50\% of the runtime on a 15 FPS stream. This exposes a critical system bottleneck where keyframe selection relies on post-hoc evaluation: the system must execute the computationally expensive dense geometric decoding process simply to determine if a frame actually contains novel geometry. Consequently, this architectural flow leads to late rejection and wasted compute on highly redundant frames.

To resolve this structural inefficiency, we introduce \framename, a lightweight feed-forward frame-gating network. \framename~evaluates incoming frames against a reference keyframe to predict a Geometric Utility Score, balancing geometric novelty against mapping cost. By moving the selection decision upstream prior to the heavy GFM feature extraction, \framename~serves as a predictive gating module that filters out uninformative frames early. Empirical evaluations on standard datasets, including TUM-RGBD \cite{sturm12iros}, 7-Scenes \cite{glocker2013real}, and EuRoC \cite{Burri25012016}, demonstrate that \framename~bypasses over 90\% of input frames while preserving the fidelity of tracking and camera pose estimation.

The primary contributions of this work are summarized as follows:
\begin{itemize}
    \item We identify the late-rejection computational bottleneck in monocular SLAM utilizing GFM front-ends, demonstrating that the primary compute cost originates from processing temporally redundant dense streams.
    \item We formalize a pairwise geometric utility score and develop a lightweight feed-forward gating network, trained via distillation from a dense teacher model, to predict frame value prior to heavy geometric decoding.
    \item We conduct extensive evaluations across multiple SLAM benchmarks, demonstrating that \framename~accelerates end-to-end system throughput by $5\times$ and reduces tracking FLOPs by over 85\% without compromising tracking accuracy.
\end{itemize}
\vspace{-1mm}
\section{Related Work}
\subsection{Visual SLAM: From Geometry to Deep Learning}
The development of Visual SLAM relies heavily on standard benchmarks such as TUM RGB-D~\cite{sturm12iros}, KITTI~\cite{geiger2013vision}, and ETH3D~\cite{schops2019bad}, which build upon earlier evaluation efforts like SLAMBench~\cite{nardi2015introducing}. Traditional algorithms bifurcate into indirect methods minimizing reprojection error over sparse features (PTAM~\cite{klein2007parallel}, ORB-SLAM~\cite{mur2015orb}) and direct methods optimizing photometric consistency (DSO~\cite{engel2016photometrically}, SVO~\cite{forster2014svo}, LSD-SLAM~\cite{engel2014lsd}). Because these handcrafted pipelines often fail in textureless regions or under extreme motion, deep learning initially replaced individual components with neural alternatives, utilizing SuperPoint~\cite{detone2018superpoint} and D2-Net~\cite{dusmanu2019d2} for detection and description alongside SuperGlue~\cite{sarlin2020superglue} or LoFTR~\cite{sun2021loftr} for matching. Building on joint optimization approaches such as BA-Net~\cite{tang2018ba} and probabilistic optimization from DeepFactors~\cite{czarnowski2020deepfactors}, DROID-SLAM integrated a differentiable Dense Bundle Adjustment layer within a recurrent framework to couple learned features with geometric solvers.

\subsection{Foundation Models and Feed-forward Reconstruction}
Geometric Foundation Models treat reconstruction as a dense regression task. Departing from incremental triangulation, DUSt3R~\cite{wang2024dust3r} introduces a feed-forward ViT architecture inspired by RAFT~\cite{teed2020raft} all-pairs correlation to regress 3D pointmaps directly from uncalibrated image pairs, implicitly inferring camera intrinsics and extrinsic poses without parametric models. This surpasses grid-based matching like GMS~\cite{bian2017gms} and achieves competitive performance against kernel-based methods like DKM~\cite{edstedt2023dkm}. Unlike MVS frameworks like MVSNet~\cite{yao2018mvsnet} requiring known poses and intrinsics for cost volumes, these models infer geometry and parameters directly. MASt3R~\cite{leroy2024grounding} extends this by learning local features alongside geometric regression. It projects dense correspondence lines from methods like PDC-Net+~\cite{truong2023pdc} into 3D space with features akin to ASLFeat~\cite{Luo2020ASLFeat}. This 3D-centric matching outperforms classical 2D pipelines, surpasses modern matchers like LightGlue~\cite{Lindenberger2023LightGlue}, and replaces learned detectors like KeyNet~\cite{keynet2019}. Although VGGT~\cite{wang2025vggt} offers strong priors using efficient attention akin to SegFormer~\cite{Xie2021SegFormer}, high memory requirements restrict it to short sequences, precluding use in long-term navigation and mapping.

\subsection{SLAM and SfM in the Foundation Model Era}

Scaling feed-forward priors for long-range trajectories remains an active challenge. MASt3R-SLAM~\cite{murai2024_mast3rslam} fuses two-view priors into a globally consistent system using pointmap matching and second-order optimization for calibration-free operation on unconstrained video. These GFM pipelines benchmark against COLMAP~\cite{schonberger2016structure} and complement neural representations like iMAP~\cite{sucar2021imap} and NICE-SLAM~\cite{zhu2022nice}. Systems also explore explicit structures like EC3R~\cite{hu2025ec3rslamefficientconsistentmonocular}, 3D Gaussian Splatting~\cite{kerbl20233d} and evaluate on challenging benchmarks like LaMAR~\cite{Sarlin2022LaMAR} and map-free settings~\cite{arnold2022map}. Unlike coordinate regression methods\cite{brachmann2023accelerated} predicting 3D coordinates directly, GFM reconstructions supply multi-view geometric priors for mapping and optimization. Aligning submaps from uncalibrated priors introduces geometric challenges. VGGT-SLAM~\cite{maggio2025vggt} and VGGT-SLAM 2.0~\cite{maggio2026vggt} address projective ambiguity where uncalibrated scenes retain a 15-degree-of-freedom homography. Optimizing on the $SL(4)$ manifold corrects distortions like shear and stretch unresolved by $Sim(3)$. Transitioning to projective optimization enables metric-quality reconstruction from uncalibrated priors, connecting to multiple view geometry~\cite{hartley2003multiple} and classical bundle adjustment~\cite{triggs1999bundle}. Concurrently, emerging approaches adopt point-based neural mapping like Point-SLAM~\cite{sandstrom2023point} to bridge reconstruction with scalable localization.

\section{Analysis}
\label{sec:analysis}
\subsection{Preliminary Experiment on Redundancy Analysis}

\begin{table}[t] 
    \centering
    \footnotesize
    \setlength{\tabcolsep}{2.5pt}
    \renewcommand{\arraystretch}{1.1}
    \footnotesize
\setlength{\tabcolsep}{2.5pt}
\renewcommand{\arraystretch}{0.9}

\resizebox{\linewidth}{!}{%
\begin{tabular}{l *{9}{S[table-format=2.1]} S[table-format=1.1]}
\toprule
Method & {desk} & {360} & {dsk2} & {rpy} & {xyz} & {flor} & {plnt} & {room} & {tdy} & {Avg} \\
\midrule
ORB3\cite{campos2021orb} & \multicolumn{1}{c}{--} & 1.7 & 21.0 & \multicolumn{1}{c}{--} & 0.9 & \multicolumn{1}{c}{--} & 3.4 & \multicolumn{1}{c}{--} & \multicolumn{1}{c}{--} & \multicolumn{1}{c}{--} \\
DPV\cite{lipson2024deep} & 11.2 & 1.8 & 2.9 & 3.0 & 1.0 & 5.7 & 2.1 & 33.0 & 8.4 & 7.6 \\
DROID\cite{teed2021droid} & 11.1 & 1.8 & 4.2 & 2.6 & 1.2 & 2.1 & 1.6 & 4.9 & 4.8 & 3.8 \\
\midrule
\multicolumn{2}{l}{MASt3R-SLAM\cite{murai2024_mast3rslam}} & & & & & & & & & \\
\midrule
Full(15FPS) & 1.6 & 4.9 & 2.3 & 2.7 & 0.9 & 2.5 & 2.0 & 6.1 & 4.1 & 3.0 \\
Keyframe(0.05FPS) & 1.5 & 4.9 & 2.6 & 2.7 & 0.8 & 2.6 & 1.9 & 6.6 & 3.7 & 3.0 \\
\textit{$\Delta$ATE} & -0.1 & -0.0 & +0.3 & +0.1 & -0.1 & +0.1 & -0.1 & +0.5 & -0.4 & +0.0 \\
\bottomrule
\label{tab:ATE_comparison}
\end{tabular}%
}
    \caption{Absolute Trajectory Error (ATE in cm) on TUM-RGBD sequence fr1 comparing full-frame tracking at 15\,FPS with keyframe-only tracking; lower is better.}
    \label{tab:ate_full_vs_kf}
\end{table}

\begin{figure}[t]
    \centering
    \includegraphics[width=1.\linewidth]{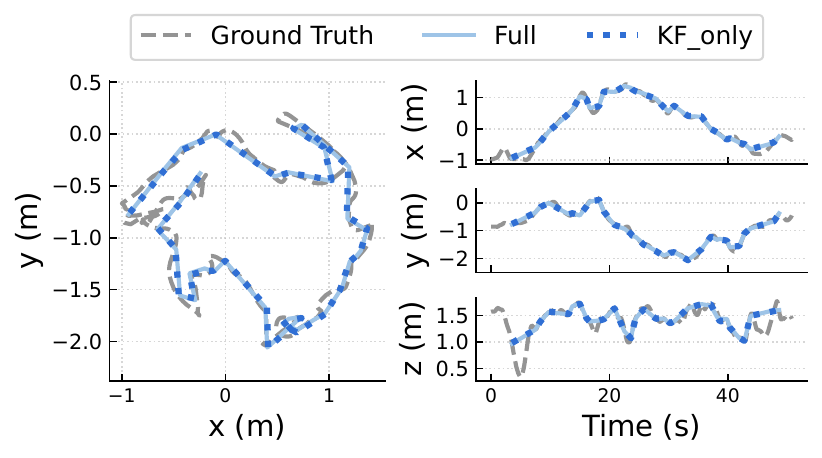}
    \caption{$Sim(3)$-aligned trajectory comparison on TUM-RGBD \textit{fr1-teddy}, contrasting full-frame (15\,FPS) tracking with keyframe-only tracking.}
    \label{fig:traj_comparison}
    \vspace{-5mm}
\end{figure}

\begin{figure}[t]
    \centering
    \includegraphics[width=.85\linewidth]{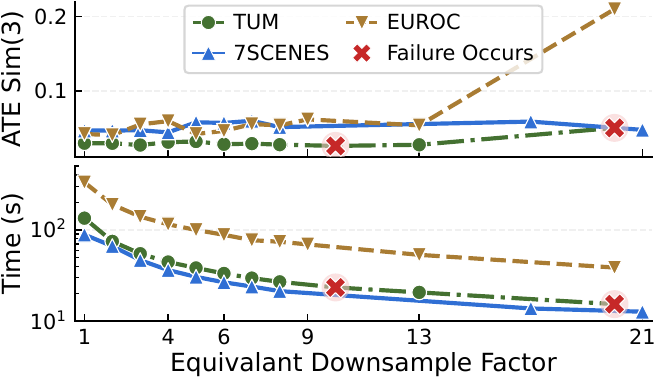}
    \caption{ATE and runtime under naive stride policies across RGB benchmarks. The shaded region marks the stride regime where TUM starts to lose scenes; red crosses denote policies evaluated on fewer scenes than the full set.}
    \label{fig:stride_analysis}
\end{figure}
To examine whether dense temporal processing is necessary in GFM-based SLAM, we conducted a preliminary study using MASt3R-SLAM as a representative framework. Specifically, we compare the standard full-frame tracking mode (15\,FPS) with a keyframe only configuration under identical backend and optimization settings.

As shown in Table~\ref{tab:ate_full_vs_kf}, using dense input and using only the keyframes selected from the same dense input yield nearly identical results. This suggests that MASt3R does not actually require a large number of supportive frames to track the motion between two keyframes. For a more intuitive illustration, we visualize one representative scene in Fig.~\ref{fig:traj_comparison}. The result shows that, in such a keyframe-based tracking paradigm, the 3D outputs of the keyframes alone are sufficient to recover stable and accurate camera poses.

\begin{figure}[ht]
    \centering
    \includegraphics[width=1.\linewidth]{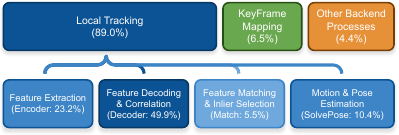}
    \caption{Averaged per-scene time breakdown profiled on TUM-RGBD without frame/model loading time. Tracker (encoder/decoder) takes most of the computation time.}
    \label{fig:time_analysis}
\end{figure}

Based on these observations, we formulate our first postulate:
\emph{Temporal redundancy in GFM-based SLAM is widespread and consistent with general RGB video characteristics.}
As shown, the system maintains trajectory integrity even with aggressive frame-skipping policies.

To further validate this, we evaluated a naive stride policy across SLAM datasets. Notably, keyframes are not uniformly distributed in time, but are instead selected by the GFM based on scene coverage. As shown in Fig.~\ref{fig:stride_analysis}, substantial differences already emerge across standard SLAM datasets such as TUM, 7Scenes, and EuRoC. The same stride can lead to tracking failure in some TUM sequences, degrade accuracy on EuRoC, while still being far from optimal on 7Scenes. Therefore, we formulate our second postulate:
\emph{There exists no universally optimal fixed stride for dense streaming. The ideal sampling frequency is intrinsically coupled with the scene's geometric complexity and motion dynamics.}

\subsection{The Paradox of Post-Inference Selection}
\label{sec:analysis_problem}

Unlike systems that update their state incrementally, such as \textit{DROID-SLAM}, \textit{MASt3R-SLAM} is built on a monolithic 3D reconstruction prior. In this paradigm, the metrics needed to assess a frame's geometric utility, such as spatial overlap or matching confidence, only become available \textbf{after} full dense decoding.

As illustrated in Fig.~\ref{fig:time_analysis}, this leads to a \textbf{computational paradox}: we must pay the full GPU price to decide if a frame is worth processing. The system attempts to prune redundant frames to ensure backend efficiency, yet it can only identify these redundancies by first squandering heavy resources on dense inference. 

Our formulation aims to break this "process-then-evaluate" cycle. By predicting geometric gain from early-stage latent features \textit{before} the dense decoder is invoked, we effectively decouple the selection decision from the inference cost. This distinguishes our work from the heuristic filtering in \textit{VGGT-SLAM} and the iterative refinement in \textit{DROID-SLAM}, providing a lean gating mechanism that ensures only informative frames reach the expensive reconstruction stage.
\section{Methodology}
\label{sec:method}

\subsection{Formalization of Geometric Utility Score}
\label{sec:metric_formalization}

We formalize the geometric utility score $\mathcal{S}$ used in the MASt3R-SLAM~\cite{murai2024_mast3rslam} keyframe selection pipeline and directly adopt it as the teacher signal for training LeanGate. Given an incoming frame $I_i$ and the latest reference keyframe $I_j$, the model predicts dense 3D pointmaps $\mathbf{P} \in \mathbb{R}^{H \times W \times 3}$, local confidence maps $C \in [0,1]^{H \times W}$, and global quality maps $Q \in \mathbb{R}^{H \times W}$. Here, $C$ measures the local certainty of a correspondence, while $Q$ reflects the spatial reliability of the predicted geometry. In the following, all quantities are defined in the valid operating regime of the MASt3R-SLAM pipeline, as used throughout our experiments.

\textit{Pixel-wise validity.} 
For each pixel $p \in \Omega_i$, the tracker establishes a correspondence $q = \mathbf{m}_{i \to j}(p)$ in $I_j$ via an iterative re-projection search ($\mathrm{IterProj}$) that optimizes the local alignment between the predicted pointmaps $\mathbf{P}_i$ and $\mathbf{P}_j$ in the aligned comparison frame. A correspondence is considered valid if it satisfies a joint three-fold constraint:
\begin{equation}
\label{eq:valid_kf}
\begin{aligned}
\mathrm{valid}_{kf}(p) \;=\;& \big(\lVert \mathbf{P}_i(p) - \mathbf{P}_j(q) \rVert_2 < \tau_{d}\big) \\
&\land\ \big(\min(C_i(p), C_j(q)) > \tau_c\big) \\
&\land\ \big(\sqrt{Q_i(p) \cdot Q_j(q)} > \tau_q\big),
\end{aligned}
\end{equation}
where $\tau_d$, $\tau_c$, and $\tau_q$ denote the thresholds for 3D distance consistency, confidence, and spatial reliability, respectively. In particular, $\tau_d$ is measured in meters.

\textit{Frame-level utility.} 
We aggregate pixel-wise validity into two complementary metrics. The \textit{matching fraction} $f_m$ measures the density of reliable constraints relative to the current frame:
\begin{equation}
\label{eq:match_frac}
f_m \;=\; \frac{1}{|\Omega_i|}\sum_{p\in\Omega_i}\mathbf{1}_{\{\mathrm{valid}_{kf}(p)\}}.
\end{equation}
To quantify geometric coverage, we define the \textit{unique fraction} $f_u$, which measures the proportion of the reference frame covered by these valid matches:
\begin{equation}
\label{eq:unique_frac}
f_u = \frac{|\mathcal{U}_{i\to j}|}{|\Omega_j|},
\end{equation}
\begin{equation}
\label{eq:unique_set}
\mathcal{U}_{i\to j}
= \left\{ q \in \Omega_j \;\middle|\;
q = \mathbf{m}_{i \to j}(p),\ 
p \in \Omega_i,\ 
\mathrm{valid}_{kf}(p)=1
\right\}.
\end{equation}
The final utility score is defined as
\begin{equation}
\mathcal{S} = \min(f_m, f_u),
\label{eq:score}
\end{equation}
which follows the original MASt3R-SLAM design and conservatively suppresses two failure modes: insufficient valid correspondences and insufficient geometric coverage. Following the MASt3R-SLAM indoor configuration, we set $\tau_d = 0.1\,\text{m}$, $\tau_c = 0.0$, and $\tau_q = 1.5$. A new keyframe is triggered whenever $\mathcal{S} < \omega_k$ (with $\omega_k = 0.33$). We directly inherit this rule from MASt3R-SLAM, and found it to be effective and robust across all datasets used in our experiments, yielding a favorable practical trade-off between tracking stability and memory efficiency.

\subsection{Feed-forward Score Regression}
\textit{From Post-hoc Assessment to Predictive Selection.}
The GFM-based SLAM systems (e.g., MASt3R-SLAM) commonly adopt a post-hoc keyframe selection paradigm: the system must first run the expensive GFMs to produce dense geometric representations and perform geometric matching, and only then can it compute utility metrics like overlap. This logic leads to substantial computation and energy waste, since frames with little mapping value still trigger nearly the same peak compute path as critical keyframes.

To reduce the overhead, we move selection upstream and formulate it as a feed-forward regression problem: we predict a geometric utility score $\tau \approx 1 - \mathcal{S}$ before entering the dense reconstruction branch. Frames predicted to have low geometric utility can skip the heavy geometry branch. To instantiate this scoring mechanism, we build a lightweight utility regressor on top of FLARE's first stage and refine the utility estimate within a single forward pass, as detailed in Sec.~\ref{sec:our_model_design}.
\subsection{The design of \framename}
\label{sec:our_model_design}

\subsubsection{Geometric Utility Score Regressor}
\label{sec:camera_latent}

Instead of explicitly injecting external pose priors, we reuse the camera-aware mechanism learned inside the foundation model and build the utility regressor on top of it.

\noindent\textbf{Camera-latent representation.}
FLARE\cite{zhang2025flarefeedforwardgeometryappearance} is a feed-forward model for joint camera pose estimation. We reuse the learnable camera/pose-related tokens in FLARE's decoder and their update pathway.
These tokens are iteratively updated across decoder layers through the pose encoding/decoding mechanism, forming a compact latent representation of the geometric relationship for the current image pair.
This decoder-layer token update is part of FLARE's internal camera-conditioning mechanism and is distinct from our iterative utility refinement head described next.
\subsubsection{Iterative Refinement of Geometric Utility}
\label{sec:iterative_refinement}
\begin{figure*}[ht]
    \centering
    
    \includegraphics[width=0.9\textwidth]{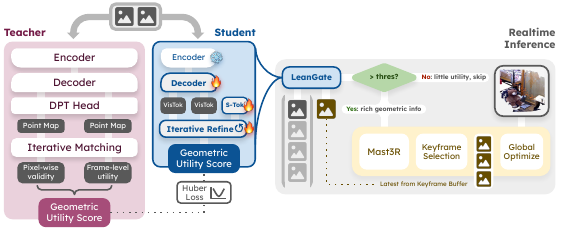}
    \caption{System Overview for \framename. Left: score-only distillation. Right: MASt3R-SLAM slimmed by \framename. The most recent keyframe (in brown color) is fed to the \framename~to calculate the score for the newly arriving frame (in grey).}
    \label{fig:system_overview}
    \vspace{-5mm}
\end{figure*}
To provide a more expressive prediction head for overlap estimation, we use an iterative overlap latent refinement head. For each input pair, we take the two decoder-extracted tokens, corresponding to the reference and current views, as the initial pairwise representation.
Conditioned on these semantic and geometric priors, a zero-initialized overlap latent is iteratively refined by a shared trunk, and the final refined latent is used for overlap prediction.

\noindent\textbf{Joint attention interaction and iterative regression.} To predict the geometric utility score, we maintain a low-dimensional latent state and instantiate a latent-conditioned \texttt{score\_token} at each refinement step.
In the regression trunk, we concatenate the \texttt{score\_token} with the pose-aware tokens and apply joint self-attention to aggregate geometric score. A lightweight prediction head then predicts an additive update to the latent utility state from the trunk output.
After several refinement steps, a readout head regresses the final utility score $\tau$ from the refined latent state.

Specifically, we design an \emph{Iterative Overlap Head} that maintains a low-dimensional latent state $\mathbf{h}_{\mathrm{ov}}\in\mathbb{R}^{8}$.
For each input pair $(I_{\mathrm{ref}}, I_{\mathrm{cur}})$, we initialize $\mathbf{h}_{\mathrm{ov}}^{(0)}=\mathbf{0}$ and perform $T$ refinement iterations on the same pairwise features, allowing the model to progressively refine its geometric score in latent space. And a system overview of \framename~is shown in Fig.~\ref{fig:system_overview}

\subsubsection{Frame Update Policy via Utility Score}
\label{sec:update_policy}
At inference time, all incoming images are handled uniformly as ordinary frames. For each frame $I_{\mathrm{cur}}$, the model predicts a utility score $\tau(I_{\mathrm{cur}})$ and applies a fixed threshold $\tau_{\mathrm{keep}}$ to decide whether the frame should be forwarded to the SLAM system. If the score passes the threshold, the frame is kept and processed by SLAM; otherwise, it is discarded immediately. Unless otherwise specified, we use $\tau_{\mathrm{keep}} = 0.5$ in all experiments.

\subsection{Geometry utility score labeling with ScanNet++}
\label{sec:data_curation}

Our goal in data curation is to convert static 3D environments into supervision that reflects SLAM-relevant geometric change.
Instead of relying on consecutive temporal clips, we construct \emph{geometric challenge pairs} to train the model to associate viewpoint variation with geometric utility score.

\subsubsection{High-fidelity labeling.}
We use ScanNet++~\cite{yeshwanth2023scannet++} as the primary data source for pseudo-label generation.
This choice is motivated by two practical considerations.
First, ScanNet++ provides high-quality reconstructions and accurate camera trajectories, which improve the reliability of correspondence-based overlap estimates.
Second, ScanNet++ lies within (or close to) the training distribution of the teacher model, MASt3R~\cite{leroy2024grounding}, which helps reduce domain mismatch during pseudo-labeling.
Unless otherwise specified, we run the teacher at an input resolution of $512\times512$ to obtain stable correspondence signals for supervision.

\subsubsection{Trajectory-agnostic sampling.}
Sequential sampling can allow a model to exploit temporal smoothness (e.g., near-constant velocity) rather than learning geometry.
To reduce this shortcut, we sample image pairs $(\mathcal{I}_i,\mathcal{I}_j)$ based on their relative camera poses, independent of temporal adjacency.
This pairwise strategy encourages the regressor to infer geometric utility from visual--spatial overlap under diverse motion patterns.
We define the supervision target as
\begin{equation}
\label{eq:tau_gt}
\tau_{\mathrm{gt}} \;=\; 1 - \mathcal{S}(\mathcal{I}_i,\mathcal{I}_j),
\end{equation}
where $\mathcal{S}(\cdot)$ is computed from MASt3R-SLAM's keyframe selection mechanism in Eq.~\ref{eq:score}. 

\subsection{Distillation and Inference Logic}
\label{sec:distillation}

This section describes how we transfer the teacher's geometric scoring behavior into a lightweight student suitable for real-time gating.

\subsubsection{Score-only distillation.}
We adopt a score-only distillation scheme: the student is trained to match the final score $\tau$, without mimicking intermediate dense features or camera intrinsics.
This design keeps training lightweight and avoids coupling the student to teacher-specific internal representations or geometric calculations.

\subsubsection{Robust regression with Huber loss.}
Pseudo-labels may still be noisy in visually challenging regions (e.g., strong illumination changes or weak texture).
We therefore use Huber loss for regression:
\begin{equation}
\label{eq:huber_obj}
\mathcal{L} \;=\; \frac{1}{N} \sum_{n=1}^{N} \ell_{\delta}\!\left(\tau_{\mathrm{stu}}^{(n)} - \tau_{\mathrm{tea}}^{(n)}\right),
\end{equation}
with
\begin{equation}
\label{eq:huber_def}
\ell_{\delta}(a)=
\begin{cases}
\frac{1}{2}a^2, & |a|\le\delta,\\
\delta\left(|a|-\frac{1}{2}\delta\right), & \text{otherwise.}
\end{cases}
\end{equation}
Huber loss behaves quadratically near zero (for precise fitting) while limiting the influence of outliers.

\subsubsection{Inference-time efficiency.}
The student model operates independently of the teacher’s dense 3D pipeline, significantly reducing computational overhead. Similar to MASt3R, it further optimizes GPU utilization by leveraging cached frames and batch inference for system-level gating.

\section{Experiment}
\label{sec:exp}
\begin{table*}[ht]
    \centering
    \small
    \caption{Reconstruction quality under different downsampling strategies on TUM RGB-D, EuRoC MAV, and 7-Scenes. We compare \framename ~against uniform striding; metrics include Completion (Comp), Chamfer distance, and F-score at 2\,cm and 5\,cm (higher is better for F-score, lower is better for distances). $\Delta$\% denotes change relative to the $1\times$ baseline; \framename ~preserves quality with $16\times$--$32\times$ fewer frames, approaching the $2\times$ stride baseline.}
    \label{tab:3dreconstruction}
    \begingroup
\setlength{\tabcolsep}{2.4pt}
\renewcommand{\arraystretch}{1.0}
\begin{tabular*}{\linewidth}{@{\extracolsep{\fill}}c l c
    S[table-format=1.3]
    S[table-format=1.3]
    S[table-format=1.3]
    S[table-format=1.3]@{}}
    \toprule
    Dataset & Method & Downsample 
    & \multicolumn{1}{c}{Comp $\downarrow$ ($\Delta\%\uparrow$)} 
    & \multicolumn{1}{c}{Chamfer $\downarrow$ ($\Delta\%\uparrow$)} 
    & \multicolumn{1}{c}{F@2cm $\uparrow$ ($\Delta\%\uparrow$)} 
    & \multicolumn{1}{c}{F@5cm $\uparrow$ ($\Delta\%\uparrow$)} \\
    \midrule

    \multirow{6}{*}{\textbf{TUM RGB-D}} 
        & DROID-SLAM~\cite{teed2021droid} & $1\times$ & 0.631 & 0.355 & 0.076 & 0.184 \\
        \cmidrule(lr){2-7}
        & MASt3R-SLAM~\cite{murai2024_mast3rslam} & & & & & \\
        & \quad All        & $1\times$  & 0.107 & 0.143 & 0.204 & 0.425 \\
        & \quad Stride 2   & $2\times$  & \multicolumn{1}{r}{0.129 ($-20.6$)}  & \multicolumn{1}{r}{0.145 ($-1.4$)}   & \multicolumn{1}{r}{0.212 (+3.9)}  & \multicolumn{1}{r}{0.433 (+1.9)} \\
        & \quad Stride 15  & $15\times$ & \multicolumn{1}{r}{0.545 ($-409.3$)} & \multicolumn{1}{r}{0.393 ($-174.8$)} & \multicolumn{1}{r}{0.188 ($-7.8$)} & \multicolumn{1}{r}{0.368 ($-13.4$)} \\
        & \quad \framename  & $16\times$ & \multicolumn{1}{r}{0.160 ($-49.5$)}  & \multicolumn{1}{r}{0.149 ($-4.2$)}   & \multicolumn{1}{r}{0.202 ($-1.0$)} & \multicolumn{1}{r}{0.422 ($-0.7$)} \\
    \midrule

    \multirow{6}{*}{\textbf{EuRoC MAV}} 
        & DROID-SLAM~\cite{teed2021droid} & $2\times$ & 1.257 & 0.705 & 0.023 & 0.133 \\
        \cmidrule(lr){2-7}
        & MASt3R-SLAM~\cite{murai2024_mast3rslam} & & & & & \\
        & \quad All        & $1\times$  & 0.271 & 0.274 & 0.030 & 0.221 \\
        & \quad Stride 2   & $2\times$  & \multicolumn{1}{r}{0.272 ($-0.4$)}   & \multicolumn{1}{r}{0.272 (+0.7)}   & \multicolumn{1}{r}{0.031 (+3.3)} & \multicolumn{1}{r}{0.224 (+1.4)} \\
        & \quad Stride 15  & $15\times$ & \multicolumn{1}{r}{0.370 ($-36.5$)}  & \multicolumn{1}{r}{0.316 ($-15.3$)} & \multicolumn{1}{r}{0.030 (+0.0)} & \multicolumn{1}{r}{0.206 ($-6.8$)} \\
        & \quad \framename  & $18\times$ & \multicolumn{1}{r}{0.348 ($-28.4$)}  & \multicolumn{1}{r}{0.298 ($-8.8$)}  & \multicolumn{1}{r}{0.031 (+3.3)} & \multicolumn{1}{r}{0.219 ($-0.9$)} \\
    \midrule

    \multirow{6}{*}{\textbf{7-Scenes}} 
        & DROID-SLAM~\cite{teed2021droid} & $2\times$ & 0.401 & 0.237 & 0.170 & 0.385 \\
        \cmidrule(lr){2-7}
        & MASt3R-SLAM~\cite{murai2024_mast3rslam} & & & & & \\
        & \quad All        & $1\times$  & 0.149 & 0.144 & 0.263 & 0.476 \\
        & \quad Stride 2   & $2\times$  & \multicolumn{1}{r}{0.150 ($-0.7$)}   & \multicolumn{1}{r}{0.143 (+0.7)}   & \multicolumn{1}{r}{0.272 (+3.4)} & \multicolumn{1}{r}{0.483 (+1.5)} \\
        & \quad Stride 15  & $15\times$ & \multicolumn{1}{r}{0.157 ($-5.4$)}   & \multicolumn{1}{r}{0.143 (+0.7)}   & \multicolumn{1}{r}{0.257 ($-2.3$)} & \multicolumn{1}{r}{0.470 ($-1.3$)} \\
        & \quad \framename  & $32\times$ & \multicolumn{1}{r}{0.140 (+6.0)}      & \multicolumn{1}{r}{0.141 (+2.1)}   & \multicolumn{1}{r}{0.264 (+0.4)} & \multicolumn{1}{r}{0.493 (+3.6)} \\
    \bottomrule
\end{tabular*}
\endgroup
\end{table*}
\begin{table*}[ht]
    \centering
    \small
    \caption{Core results across datasets comparing trajectory accuracy and computational efficiency. For Droid-SLAM, we report both single-GPU and parallel-mode profiling on the same device using official settings.}
    \label{tab:core_results}
    \begingroup
\setlength{\tabcolsep}{2.4pt}
\renewcommand{\arraystretch}{1.0}
\begin{tabular*}{\linewidth}{@{\extracolsep{\fill}}c l c S[table-format=1.2] S[table-format=3.2] S[table-format=5.2] S[table-format=5.2] S[table-format=5.2]@{}}
\toprule
\multirow{3}{*}{Dataset} & \multirow{3}{*}{Model}
& \multirow{3}{*}{Downsample}
& \multicolumn{1}{c}{\multirow{3}{*}{ATE [cm] $\downarrow$}}
& \multicolumn{1}{c}{\multirow{3}{*}{\shortstack{Time [s] $\downarrow$}}}
& \multicolumn{3}{c}{\shortstack{Calculations [TFLOPs] $\downarrow$}} \\
\cmidrule(l{4pt}r{4pt}){6-8}
&  &  &  &  & \multicolumn{1}{c}{Frame Select} & \multicolumn{1}{c}{SLAM} & \multicolumn{1}{c}{Total} \\
\midrule
\multirow{4}{*}{\textbf{TUM RGB-D}}
& DPV-SLAM~\cite{lipson2024deep}
& $1\times$ & 7.6 & 43.79 & \multicolumn{1}{c}{---} & \multicolumn{1}{c}{---} & \multicolumn{1}{c}{---} \\
& DROID-SLAM~\cite{teed2021droid}
& $1\times$ & 3.8 & \multicolumn{1}{r}{41.78/39.56} & \multicolumn{1}{c}{---} & \multicolumn{1}{c}{---} & \multicolumn{1}{c}{---} \\
& MASt3R-SLAM~\cite{murai2024_mast3rslam}
& $2\times$ & 3.00 & 74.95 & \multicolumn{1}{c}{---} & 6698.55 & 6698.55 \\
& \framename
& $15.58\times$ & 2.56 & 18.18 & 532.05 & 461.41 & 993.46 \\
\midrule
\multirow{4}{*}{\textbf{EuRoC MAV}}
& DPV-SLAM~\cite{lipson2024deep}
& $2\times$ & 2.4 & 122.15 & \multicolumn{1}{c}{---} & \multicolumn{1}{c}{---} & \multicolumn{1}{c}{---} \\
& DROID-SLAM~\cite{teed2021droid}
& $2\times$ & 2.2 & \multicolumn{1}{r}{127.45/103.68} & \multicolumn{1}{c}{---} & \multicolumn{1}{c}{---} & \multicolumn{1}{c}{---} \\
& MASt3R-SLAM~\cite{murai2024_mast3rslam}
& $2\times$ & 4.09 & 189.50 & \multicolumn{1}{c}{---} & 20835.09 & 20835.09 \\
& \framename
& $18.60\times$ & 4.90 & 44.63 & 1592.16 & 1206.14 & 2798.31 \\
\midrule
\multirow{4}{*}{\textbf{7-Scenes}}
& DPV-SLAM~\cite{lipson2024deep}
& $2\times$ & 5.4 & 38.44 & \multicolumn{1}{c}{---} & \multicolumn{1}{c}{---} & \multicolumn{1}{c}{---} \\
& DROID-SLAM~\cite{teed2021droid}
& $2\times$ & 4.9 & \multicolumn{1}{r}{41.29/40.75} & \multicolumn{1}{c}{---} & \multicolumn{1}{c}{---} & \multicolumn{1}{c}{---} \\
& MASt3R-SLAM~\cite{murai2024_mast3rslam}
& $2\times$ & 4.71 & 66.46 & \multicolumn{1}{c}{---} & 4978.32 & 4978.32 \\
& \framename
& $32.26\times$ & 4.61 & 12.64 & 318.75 & 174.41 & 493.16 \\
\bottomrule
\end{tabular*}
\endgroup
\end{table*}
\vspace{-3mm}

\subsection{Experiment Setup}
\label{sec:exp_setup}

\subsubsection{Teacher-led Dataset Generation}
\label{sec:teacher_dataset}

We construct a large-scale supervision dataset from 150 ScanNet++ scenes.
To balance geometric diversity and pair density, we downsample the original 60\,FPS iPhone streams to 12\,FPS, yielding $0.5$\,M training pairs.
We use an 80/20 train/eval split on scenes.

For each pair $(\mathcal{I}_i,\mathcal{I}_j)$, we compute the ground-truth utility score $\tau_{\mathrm{gt}}$ using MASt3R-SLAM, following Eq.~\ref{eq:tau_gt}.
The resulting labels are approximately bell-shaped, covering a broad range from redundant pairs to visually novel viewpoints.

\subsubsection{Implementation and Training}
\label{sec:impl_training}

Our iterative utility regressor builds on the pretrained FLARE\cite{zhang2025flare} backbone.
For faster inference, we truncate the decoder from 12 to 6 layers.
We re-use the selected decoder blocks and camera-conditioning modules to keep the original geometric prior and stable feature scaling.
The regression head maintains an 8-D latent state and performs $K{=}4$ refinement iterations to output a scalar geometric utility score.

\textit{Training details.}
We supervise the model with Huber loss ($\delta{=}0.1$) for robustness to teacher labels.
Training runs for 20 epochs with AdamW and decoupled learning rates:
\emph{Information Score Modules} $3\times10^{-4}$,
\emph{Decoder} $5\times10^{-5}$.

All experiments are conducted in a Docker environment (CUDA~12.4, Python~3.11) on a node with $8\times$ NVIDIA RTX A5000 (24\,GB) GPUs.
We use Distributed Data Parallel (DDP) with total batch size of 1024, BF16 mixed precision, and a 5-epoch linear warmup.
Input images are resized to $256\times256$ to balance geometric detail and throughput.
\subsection{3D Reconstruction}
\label{sec:recon}
We evaluate reconstruction quality using standard point-to-point metrics in Table~\ref{tab:3dreconstruction}. Accuracy (Acc) measures the mean nearest-neighbor distance from predicted points to the reference surface (Pred$\rightarrow$Ref), while Completeness (Comp) measures the reverse direction (Ref$\rightarrow$Pred). Chamfer-L1 is defined as $0.5(\text{Acc}+\text{Comp})$. We also report F-scores at 2\,cm and 5\,cm, where higher values indicate better reconstruction quality.
All methods are evaluated in \emph{calibrated} mode for a unified comparison protocol. Although this setting may slightly underestimate performance in some cases, it ensures consistency across methods.

Since \textit{TUM-RGBD}, \textit{EuRoC}, and \textit{7-Scenes} do not provide consistent dense 3D ground truth across sequences, and depth or stereo observations are not uniformly available under a shared dense reconstruction setup, standard alternatives such as TSDF are not directly applicable in a unified evaluation.

To enable consistent comparison, we use dense reconstructions generated by Map-Anything~\cite{keetha2026mapanything} as a \emph{proxy reference geometry}. We intentionally avoid using dense MASt3R as reference to reduce bias toward methods with similar model priors. Therefore, the reported metrics reflect geometric consistency with respect to a common external proxy, rather than absolute accuracy against true 3D ground truth.

As shown in Table~\ref{tab:3dreconstruction}, \framename\ consistently provides a stronger efficiency--quality trade-off than naive stride-based subsampling under aggressive frame reduction. On \textit{TUM-RGBD} and \textit{EuRoC}, it remains substantially closer to the full-frame reconstruction while clearly outperforming the high-stride baseline in both geometric distance and F-score. On \textit{7-Scenes}, \framename\ even surpasses the full-frame setting, suggesting that our frame selection strategy can remove redundant views while preserving, and in some cases improving, reconstruction fidelity.
\subsection{SLAM performance.}

\begin{figure}[htbp]
    \centering
    \small
    \vspace{-0.5\baselineskip}
    \includegraphics[width=0.4\textwidth]{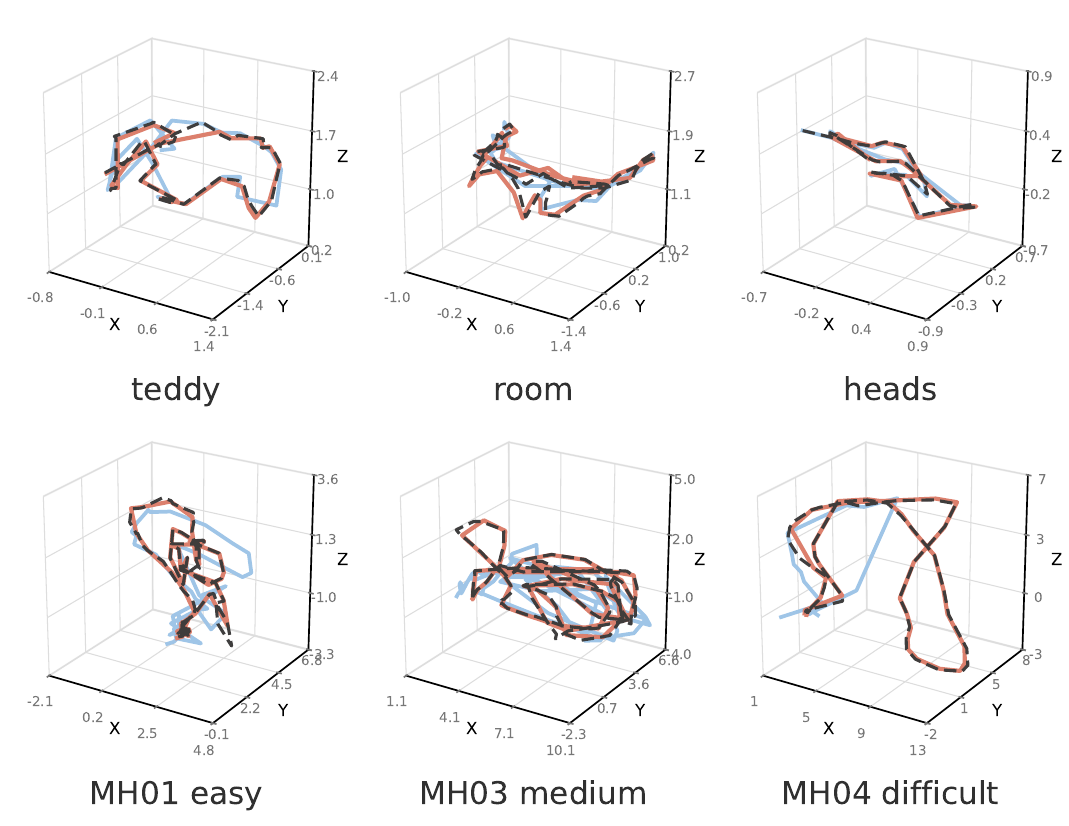}
    \vspace{-0.3\baselineskip}
    \caption{Qualitative 3D trajectory comparisons on TUM-RGBD (fr1-teddy, fr1-room), 7-Scenes (heads-seq01), and EuRoC (MH01 easy, MH03 medium, MH04 difficult). Black denotes ground truth; orange shows Sim(3)-aligned estimates from \framename-filtered RGB streams; blue shows Sim(3)-aligned stride-15 results selected to match a similar retained-frame budget, illustrating our method's tracking consistency under aggressive frame pruning.}
    \label{fig:qualitative_3d_traj}
\end{figure}

\textbf{Efficiency vs. Accuracy.} We evaluated \framename~with MASt3R-SLAM~\cite{murai2024_mast3rslam} under identical single-threaded monocular settings. Our system achieves a $4.1\times$--$5.3\times$ end-to-end speedup by pruning over 90\% redundant frames as shown in Table~\ref{tab:core_results}. While the accuracy remains close to the baseline in most scenes, we observe small degradation in EuRoC sequences, highlighting the need for future studies on robustness at grayscale images. Additionally, we visualized the 3D trajectories for several scenes in Fig.~\ref{fig:qualitative_3d_traj}. Compared with the stride-based method, our trajectories preserve finer details and adhere more closely to the true motion paths. 

In standalone evaluations of SLAM tracking costs, \framename~delivers a $5\times$--$10\times$ boost in tracking speed. This gain stems from our decoupled architecture, which enables the tracking stream to run in parallel. Furthermore, we find that \framename~does not yet fully utilize high-end GPU resources, suggesting additional headroom for throughput scaling.

We also report FLOP counts. These are not identical to actual CUDA operation counts because MASt3R-SLAM includes backend optimizations with custom CUDA kernels that lack a standard profiling strategy.
In the table, we report the best available estimates from \textit{fvcore} and \textit{pytorch profiling}. 
\begin{table}[ht]
\centering
\caption{Ablation on the ScanNet++ validation set for utility-score regression. The model design block evaluates architecture choices (iterative head and decoder depth), and the model training evaluates optimization choices (decoder freezing and pre-training). We report MAE and RMSE (lower is better); bold indicates the best value in each metric.}
\label{tab:ablation}
\setlength{\tabcolsep}{3.5pt}
\renewcommand{\arraystretch}{1.0}
\small
\begin{tabular}{llS[table-format=1.4]S[table-format=1.4]}
\toprule
Category & Configuration & {MAE $\downarrow$} & {RMSE $\downarrow$} \\
\midrule
\multicolumn{4}{l}{\textbf{Model Design}} \\
\cmidrule(lr){1-4}
Default & baseline (dec6) & \multicolumn{1}{c}{\textbf{0.0281}} & \multicolumn{1}{c}{\textbf{0.0400}} \\
Head Design & w/o iterative head & 0.0352 & 0.0488 \\
Decoder Depth & dec6 & 0.0303 & 0.0430 \\
Decoder Depth & dec3 & 0.0347 & 0.0492 \\
\midrule
\multicolumn{4}{l}{\textbf{Model Training}} \\
\cmidrule(lr){1-4}
Decoder Freeze & dec12 frozen & 0.0343 & 0.0488 \\
Pre-training & dec3 random init & 0.0537 & 0.0765 \\
Pre-training & dec6 random init & 0.0635 & 0.0883 \\
\bottomrule
\end{tabular}
\vspace{-3mm}
\end{table}

\subsection{Ablation Study}
\label{sec:ablation}

To investigate the individual contributions of our proposed components, we conduct extensive ablation studies on the ScanNet++ validation set, covering both architectural design and training strategies.

We ablate architectural choices, focusing on the scoring head and decoder depth. Table~\ref{tab:ablation} shows that the scoring head dominates performance, supporting the view that iterative refinement can replace explicit geometric matching. Increasing decoder depth from \textit{dec3} (first 3 layers) to \textit{dec6} (first 6 layers) consistently improves final precision, suggesting that decoder-stage cross-attention is crucial for effective feature aggregation. Since iteration count is not the inference bottleneck, we set \textit{iter}=4 (as in FLARE) to maximize accuracy.

We next study optimization choices. Table~\ref{tab:ablation} shows that pre-training is the dominant factor: training from random initialization leads to a substantial degradation in performance, suggesting that pre-training yields strong geometric priors that transfer to downstream refinement.
\vspace{-3mm}
\section{Conclusion}
\label{sec:conclusion}

In this paper, we introduced \framename, a lightweight feed-forward framework designed to quantify and predict frame-level information density for GFM-based systems. 
By identifying that high-fidelity processing is not uniformly required across all temporal observations, our work addresses a critical bottleneck in spatial tasks: the inherent computational redundancy inherent to the deployment of massive GFMs for downstream tasks such as SLAM and 3D reconstruction. 

Trained on $0.5$\,M samples from ScanNet++, the model already achieves strong performance across diverse indoor environments. We anticipate that further scaling of the training data will allow it to fully decouple from specific pre-training-weight dependencies and support robust acceleration for both indoor and outdoor monocular SLAM.

\bibliographystyle{IEEEtran}
\bibliography{referenece}
\end{document}